\documentclass[conference]{IEEEtran}
\IEEEoverridecommandlockouts
\usepackage{cite}
\usepackage{amsmath,amssymb,amsfonts}
\usepackage{graphicx}
\usepackage{textcomp}
\usepackage{xcolor}
\usepackage{filecontents}
\usepackage{multirow}
\usepackage{array, tabularx,makecell}
\usepackage{algorithm}
\usepackage{algpseudocode}
\usepackage{mathtools,nccmath}
\usepackage{lipsum,adjustbox}
\usepackage{pgfplots}
\usepackage{pgfplotstable}
\usepackage{siunitx}
\usepackage{subcaption}
\usepackage{caption}
\usepackage{pgfplots}

\usepgfplotslibrary{groupplots}

\newlength\mylen

\DeclarePairedDelimiter\abs{\lvert}{\rvert}%
\def\BibTeX{{\rm B\kern-.05em{\sc i\kern-.025em b}\kern-.08em
    T\kern-.1667em\lower.7ex\hbox{E}\kern-.125emX}}
\pgfplotsset{compat=1.14}
\begin{document}
\newcommand{\mysup}[1]{\textsuperscript{\raisebox{.2pt}{#1}}}

\title{Improving Prediction Accuracy in Building Performance Models Using Generative Adversarial Networks (GANs)}

\author
{\IEEEauthorblockN{Chanachok Chokwitthaya\mysup{\mbox{*}}, Edward Collier\mysup{$\dagger$}, Yimin Zhu\mysup{\mbox{*}}, Supratik Mukhopadhyay\mysup{$\dagger$}}
\IEEEauthorblockA{\textit{\mysup{\mbox{*}}Department of Construction Management and  \mysup{$\dagger$}Department of Computer Science} \\
\textit{Louisiana State University}\\
Baton Rouge, USA \\
Email: \{cchokw1, ecoll28, yiminzhu, supratik\} @lsu.edu}

}

\maketitle

\begin{abstract}
Building performance discrepancies between building design and operation are one of the causes that lead many new designs fail to achieve their goals and objectives. One of main factors contributing to the discrepancy is occupant behaviors. Occupants responding to a new design are influenced by several factors. Existing building performance models (BPMs) ignore or partially address those factors (called contextual factors) while developing BPMs. To potentially reduce the discrepancies and improve the prediction accuracy of BPMs, this paper proposes a computational framework for learning mixture models by using Generative Adversarial Networks (GANs) that appropriately combining existing BPMs with knowledge on occupant behaviors to contextual factors in new designs.  Immersive virtual environments (IVEs) experiments are used to acquire data on such behaviors.  Performance targets are used to guide appropriate combination of existing BPMs with knowledge on occupant behaviors. The resulting model obtained is called an augmented BPM.  Two different experiments related to occupants’ lighting behaviors are shown as case study. The results reveal that augmented BPMs significantly outperformed existing BPMs with respect to achieving specified performance targets. The case study confirms the potential of the computational framework for improving prediction accuracy of BPMs during design.

\end{abstract}

\begin{IEEEkeywords}
occupant behavior, mixture model, building performance model, generative adversarial network, immersive virtual reality
\end{IEEEkeywords}

\section{Introduction}
Building designs define characteristics, functions,  and contexts of buildings according to objectives and goals of a building project. Building performance is an important component during designs that needs designer’s attention. It reflects how well buildings perform regarding to many components such as energy, occupants’ comforts, and control systems. Building performance models (BPMs) are tools that support designers to investigate, predict, and understand the performance of buildings and make decisions during design. Several BPMs are used to optimize building performance during design, e.g., BPMs for predicting energy consumption (electricity consumption), BPMs for predicting building performance (heat loss and air quality), and BPMs for predicting occupants’ interactions with building components (light switches, blinds, and windows). For instance, designers use lighting BPMs to estimate occupants’ light switch behaviors. 
Empirical evidences have shown the existence of significant performance discrepancies between predictions during design and the actual performance of building operations \cite{van2016review, frei2017performance}. The performance discrepancies may contribute to undesired buildings’ performance such as unexpected energy consumption, building degradation, and occupants’ discomfort. Many factors may contribute to the discrepancies.  Occupant behaviors are one of the crucial contributing factors since they are uncertain, complex, and difficult to understand and model \cite{hong2016occupant}. Moreover, they may be influenced by many factors such as one’s sense of control, building characteristics, building services systems and operations, and climates, which make it challenging to accurately capture them while developing BPMs \cite{o2014contextual}.

Most BPMs are mathematically developed by finding the relationships between dependent and independent variables of interest. Generally, traditional methods, questionnaires \cite{attia2012development,feng2016preliminary} and field studies \cite{santin2009effect,andersen2011modelling}, are used to collect occupant behavior data (dependent variables) with respect to environmental factors (independent variables). For instance, Hunt \cite{hunt1980predicting} used field study to observe occupants’ lighting behaviors in an existing building for almost a year. He used minimum working area illuminance as a predictor to predict whether occupants switch the light on. The main advantage of using traditional methods in acquiring data on occupant behaviors is that a large pool of continuous data can be obtained, which is suitable for developing BPMs. However, capabilities of traditional research methods for studying occupant’ behavior are limited in many aspects. First, such data only represent occupant behaviors in existing buildings. Contexts of existing buildings may differ from those of new designs, which may influence occupant behaviors differently. Second, since the data of occupant behaviors are obtained from existing buildings, some factors that influence occupant behaviors in new designs may not be captured (such as contextual factors). Contextual factors are generally ignored or partially addressed in existing BPMs. These shortcomings cause reduction in the predictive capability of existing BPMs that in turn gives rise to performance gaps between predictions generated during design and actual buildings.    

IVEs can be alternative tools to support occupant behavior data collections. They are rich multisensory computer simulations that can mentally immerse users in the simulations. IVEs have been used in several research areas such as emergency situations \cite{kinateder2014social,kobes2010exit}, driving behaviors \cite{osman2015impact,rumschlag2015effects}, and building designs \cite{heydarian2014immersive,niu2016virtual}. IVEs have been proven to be capable of simulating physical environments, providing senses of reality, and capturing users’ responses. 

The paper proposes a computational framework to reduce performance discrepancy between predictions made during   design and actual building operation by combining knowledge about occupant behaviors responding to contextual and design-specific factors of new buildings with  existing BPMs. IVEs are used as tools to capture occupant behaviors. The framework uses Generative Adversarial Networks (GANs) for learning mixture models that enable appropriately combining existing BPMs with knowledge of occupant behaviors obtained from IVE to produce augmented BPMs with improved predictive power. Performance targets are used as guides to achieve appropriate combination. The computational framework offers a novel approach for improving the prediction accuracy of BPMs during design and reduce the performance discrepancy between predictions and the actual operations.

The contributions of this paper is:

\begin{itemize}
	\item We offer a computational framework to combine existing BPMs with knowledge of occupant behaviors responding to contextual factors of new building designs obtained from IVE experiments. The work contributes to the development of a novel approach for minimizing the discrepancy in building performances between predictions during designs and the actual performance during building operations, and thus allowing improved future building designs.  
\end{itemize}

\section{Related Work}
\subsection{Building Performance Models (BPMs)}
A lot of research have been devoted to developing techniques for creating BPMs. Examples of how researchers develop and use BPMs are summarized as follows.

Hunt \cite{hunt1980predicting} developed a BPM for predicting manual lighting control based on a switch-on probability and minimum working area illuminance. The BPM was developed by using field study data where sensors were installed in experimental offices to capture occupant interaction with artificial light switches.  The BPM was expressed in terms of a statistical Probit model. Likewise, Nicol \cite{nicol2001characterising} developed BPMs to predict occupant windows, lighting, blinds, heaters, and fans usages based on outdoor temperature in naturally ventilated buildings from survey data. Probit analysis was used to determine the dependence of  occupant buildings’ usages on outdoor temperature. Newsham \cite{newsham1994manual} developed and improved a computer-based thermal model “FENESTRA” by providing an algorithm to simulate manual blind operation with respect to light switching described by Hunt’s model. From the results of his model, he suggested that occupant behavior may significantly influence predictions of thermal energy consumption. Reinhart \cite{reinhart2004lightswitch} proposed an algorithm  called “Lightswitch-2002” to estimate electric lighting energy demand of light switches. It was integrated into many simulation programs, such as design support tool (Lightswitch Wizard \cite{reinhart2003lightswitch}), and whole building energy simulation tool (ESP-R \cite{bourgeois2004adding}). The algorithm included an occupancy model, which considered profiles of occupants and minimum working area illuminance similar to Hunt’s approach, and a dynamic daylight simulation to predict electric lighting demand. The algorithm considered daytime switch-on proability in addition to probability of switching the light on upon arrival. Similary, Gunay et al. \cite{gunay2017development} formulated BPMs for adaptive lighting and blinds controls algorithm. Their BPMs include concurrent solar irradiance as an additional predictor for occupant lighting preferences, beside minimum working area illuminance and intermediate occupancy in other works.

Traditionally, BPMs are developed based on data acquired using occupant behavior study approaches, namely questionnaires and field study. Most of the existing BPMs are in form of the correlation between independent variables (environments and buildings’ design-specific factors) and dependent variable (occupant behaviors). The researchers illustrate the relationships by using statistical modelling such as regression models \cite{hunt1980predicting,da2013occupants}.

\subsection{Occupant Behavior Research Methods}
Questionnaires are a common method to study occupant behaviors. Questionnaires can be directed to subjects that researchers desire to investigate. They can also handle large-scale experiments. For example, Attia et al. \cite{attia2012development} used questionnaires to collect occupant behavior data related to household device usages in residential apartments in various areas in Egypt. They applied the data obtained from the questionnaires to construct benchmarks for building energy simulations. Feng et al. \cite{feng2016preliminary} used questionnaires to observe occupant behaviors related to air conditioning (AC) patterns. The information acquired from the questionnaires were used to categorize occupants’ switching on/off AC behaviors. Questionnaires are used in research on multiple aspects of interest in several places simultaneously. For instance, Nicol \cite{nicol2001characterising} studied occupant behavior on windows, lightings, blinds, heat, and fans usage by using questionnaires in the UK, Pakistan, and Europe. Even though questionnaires provide various advantages, an important disadvantage is that they are not able to quantitatively determine the correlation between the contexts and the occupant behaviors.

The field monitoring method has been used in many studies such as light switching \cite{boyce1980observations}, predicting window opening \cite{rijal2007using}, energy usage for space and water heating \cite{santin2009effect}, occupants’ heating set-point \cite{andersen2011modelling}, occupant interactions with shading and lighting \cite{sadeghi2016occupant}, and occupant plug-in equipment use \cite{gunay2016modeling}. One of the advantages of this method is that the collected data are continuous and acquiring large samples is possible since multiple sensors are deployed with long period of time. Another advantage is that the method is capable of providing quantitative relationships between the occupant behaviors and the contexts. However, this method has many limitations, including (1) normally, data are collected in time intervals, e.g. every 30 minutes, and some critical events may be missed if they occur during the intervals, (2) other equipment may interfere with sensors and distort information of occupant behaviors and contexts, (3) many assumptions with respect to occupant behaviors and design contexts such as occupant schedules, variables that drive behaviors, and purposes of occupant response to building systems have to be made to derive the BPMs.

\subsection{Immersive Virtual Environment (IVE)}
Clearly, the traditional occupant behavior research methods described above typically rely on observations of occupants in existing buildings. Since occupant behaviors are context sensitive, findings from such observations can certainly contain biases and uncertainties. Thus, applying such findings to new designs may lead to significant variances in predictions. We suggest an alternative method to study and observe occupant behaviors during building designs by using immersive virtual environments (IVEs). There are several reasons showing that IVEs are good candidates for studying and observing occupant behaviors in buildings. For instance, IVEs allow users to control confounding and isolating variables of interest, to be immersed in their settings, and to constantly maintain variables of interest during conducting experiments \cite{heydarian2015immersive}. Previous works that show the abilities of IVEs as alternative tools to study occupant behaviors are summarized as follows.

In human behaviors related studies, Heydarian et al. \cite{heydarian2015immersive} used IVEs to study occupant behaviors related to lighting and shade usages. Saeidi et al. \cite{saeidi2015measuring} evaluated data on  occupants’ lighting behaviors acquired from  IVEs and showed that IVEs were capable of replicating experiences in physical environments. A framework for integrating bulding designs with IVEs was also developed by Niu et al. \cite{niu2016virtual}. The purpose of the framework was to help building designers capture occupant preferences and identify context patterns. They concluded that integrating building designs with IVEs using their framework helped designers to understand occupant behavior and identify design contexts that guide occupants to act corresponding to design intentions. Another work of Saeidi et al. \cite{saeidi2018spatial} conducted an experiment to study occupants’ lighting preferences in IVEs and compared the resulting data with respect to that collected from physical sensors. They found good agreement between the occupant’s preferences in IVEs and those in actual  physical environments.

\subsection{Generative Advesarial Networks (GANs)}
Deep learning has grown in popularity in recent years \cite{lecun2015deep,goodfellow2014generative,Collier2,Saikat,DeepSAT}. Generative Adversarial Networks (GANs) were proposed in \cite{goodfellow2014generative}. GANs have been successfully used in various domains \cite{collier}, especially image synthesis.

Ledig et al. \cite{ledig2017photo} used GANs to learn and recover photo-realistic textures from downsampled images. They proposed super-resolution GANs (SRGANs) that can estimate photo-realistic super-resolution images with high upscaling factors. Radford et al. \cite{radford2015unsupervised} introduced deep convolutional generative adversarial networks (DCGANs) for generating realistic and high resolution images. They showed that DCGANs outperformed other unsupervised algorithms (K-means, Random Forest (RF), and Transductive Support Vector Machines (TSVMs)). Wang and Gupta \cite{wang2016generative}, introduced Style and Structure Generative Adversarial Networks (S2-GANs). S2-GANs addressed structure and style in image generation process. S2-GANs have abilities to generate  photo-realistic high-resolution images, in addition to having a more robust and stable training method compared to standard GANs. Apart from 2D image generation, Wu et al. \cite{wu2016learning} introduced 3D-GANs that were capable of generating 3D objects  by combining volumetric convolutional networks with GANs. 

From the previous works, we have seen abilities of GANs to produce synthetic images that are close to real images from arbitrary image clues (noises). Correspondingly, we use GANs to produce augmented BPMs that are close to the performance targets by combining existing BPMs with the knowledge on occupant’ behaviors responding to contextual factors in new building designs.    

\section{Framework of Mixture Model}

\begin{figure}[htbp]
	\centering
	\includegraphics[width=0.8\columnwidth]{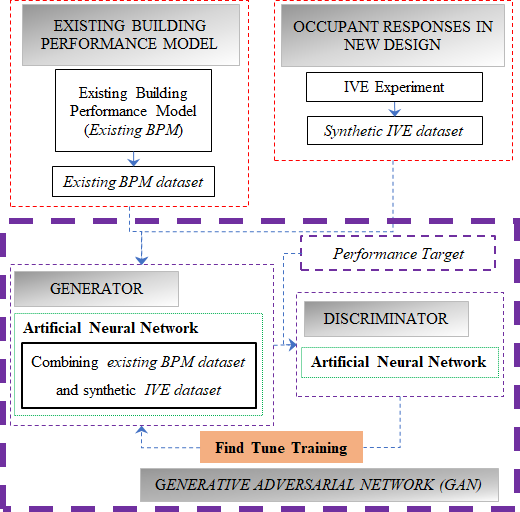}
	\caption{Framework of proposed mixture model.}
	\label{fig1}
\end{figure}

Due to the lack of ability to accurately model occupant behaviors in existing BPMs for new designs, we propose a framework to enhance BPMs by appropriately combining existing BPMs and with knowledge of occupant behaviors in new design obtained from IVE experiments (IVE datasets). There are four major components involved in the framework, namely an existing BPM, occupant responses in a new design, a performance target, and Generative Adversarial Networks (GANs). An existing BPM is a BPM that is constructed from occupant behavior data in an existing building. Occupant responses in a new design are occupant behavior data that obtained from an IVE experiment, which exposes the occupant with an environment of a new design and considers factors that are ignored in an existing BPM (contextual factors). A performance target is used as a guide for combining an existing BPM and occupant response in a new design such as building benchmarks, historically desired occupant data, and desired building performance. GANs \cite{goodfellow2014generative} are used to create  mixture models that allow appropriate combination of an existing BPM and knowledge of occupant behaviors as obtained from IVE experiments guided by a performance target (Fig.~\ref{fig1}). In the framework, we define the dataset obtained by sampling IID from an existing BPM as the existing BPM dataset. GANs comprise of two major components: a generator and a discriminator. The generator is an artificial neural network (ANN) which uses the existing BPM dataset and the IVE dataset as input and produces outputs as a mixture distribution of the existing BPM dataset and the IVE dataset (called augmented BPM). The performance predicted based on the resulting mixture distribution is intended to be as close as possible to the given performance targets. The discriminator is an ANN that tries to discriminate between the performance predictions obtained from mixture distribution generated by the generator and the performance target. During training, the generator and the discriminator engage in a minimax game between themselves where the generator intends to produce a mixture distribution so that the performance targets are met, and the discriminator tries to determine if the generator meets the performance targets. The trainings continue until a defined convergence criterion (maximum iterations, discrepancy measured between the predictions of the generator and the targets is below a threshold) is reached. Once training converges, the resulting generator obtained is the augmented BPM.

\section{Case Study}

\subsection{Existing Building Performance Models and Targets}
Two experiments related to occupant light switching behaviors are conducted. In the first experiment, a model for  predicting occupant light switching behaviors developed by \cite{hunt1980predicting} is used as the existing BPM. For the performance target, we use the probabilities of switching on as provided by a probit model described in \cite{da2013occupants}. In the second experiment, the existing BPM consisted of a model for predicting occupant light switching behaviors developed by \cite{gunay2017development}. The performance target is similar to the performance target in the first experiment. In the existing BPMs, Probit regression is used to represent the relationships between probabilities of an occupant switching on and work area illuminance as shown below: 

\newenvironment{conditions}
{\par\vspace{\abovedisplayskip}\noindent
	\begin{tabular}{>{$}l<{$} @{} >{${}}c<{{}$} @{} l}}
	{\end{tabular}\par\vspace{\belowdisplayskip}}

\begin{equation}
p = a + \frac{c}{1 + \exp(-(dm+bE_{lux}))}\label{eq1}
\end{equation}
where:
\begin{conditions}
	p     			& = &  probability of switching the light on, \\
	E_{lux}     	& = &  the working area illuminance (lux), \\   
	a, b, c, d, m 	& = &  constants  given in TABLE ~\ref{tab1}.
\end{conditions}

\begin{table}[htbp]
	\caption{Existing BPMs and Performance Targets}
	\renewcommand\arraystretch{1.0}
	\begin{center}
		\centering\begin{tabular}{|>{\centering\arraybackslash}m{1.5cm}|>{\centering\arraybackslash}m{2.5cm}|>{\centering\arraybackslash}m{2.5cm}|}
			\hline
			{} & \textbf{Experiment 1}& \textbf{Experiment 2}\\[0.05cm]
			\hline
			\multirow{6}{*}{\parbox{0.08\textwidth}{\centering \textbf{Existing BPM}}} & \multicolumn{1}{l|}{$a = -0.0175$} & \multicolumn{1}{|l|}{$a = 0$}\\[0.05cm]
			& \multicolumn{1}{l|}{$b = -4.0835$} & \multicolumn{1}{|l|}{$b = -0.005$}\\[0.05cm]
			& \multicolumn{1}{l|}{$c = 1.0361$} & \multicolumn{1}{|l|}{$c = 1$}\\[0.05cm]
			& \multicolumn{1}{l|}{$d = -4.0835$} & \multicolumn{1}{|l|}{$d = 1$}\\[0.05cm]
			& \multicolumn{1}{l|}{$m = -1.8223$} & \multicolumn{1}{|l|}{$m = -0.170$}\\[0.05cm]
			& \multicolumn{1}{l|}{$E_{lux} = log_{10}lux$} & \multicolumn{1}{|l|}{$E_{lux} = lux$}\\[0.05cm]
			\hline
			\multirow{6}{*}{\parbox{0.08\textwidth}{\centering \textbf{Performance Target}}} & \multicolumn{1}{l|}{$a = 0$} & \multicolumn{1}{|l|}{$a = 0$} \\[0.05cm]
			& \multicolumn{1}{l|}{$b = -0.003$} & \multicolumn{1}{|l|}{$b = -0.003$} \\[0.05cm]
			& \multicolumn{1}{l|}{$c = 1$} & \multicolumn{1}{|l|}{$c = 1$} \\[0.05cm]
			& \multicolumn{1}{l|}{$d = 1$} & \multicolumn{1}{|l|}{$d = 1$}\\[0.05cm]
			& \multicolumn{1}{l|}{$m = 2.035$} & \multicolumn{1}{|l|}{$m = 2.035$}\\[0.05cm]
			& \multicolumn{1}{l|}{$E_{lux} = log_{10}lux$} & \multicolumn{1}{|l|}{$E_{lux} = lux$} \\[0.05cm]
			\hline
		\end{tabular}
		\label{tab1}
	\end{center}
\end{table}

Independent and identically distributed (IID) samples of the existing BPMs and the performance targets are generated by using Monte Carlo simulations. Data of $E_{lux}$ are randomly sampled using a normal distribution. The data are taken as inputs to compute outputs (probability of switching on ($p$)) by using \eqref{eq1}. Data of $p$ and $E_{lux}$ are used in the computational framework.  

\subsection{Occupant Light Switching Behaviors in New Design}

\begin{figure}[htbp]
	\centerline{\includegraphics[width=0.7\columnwidth]{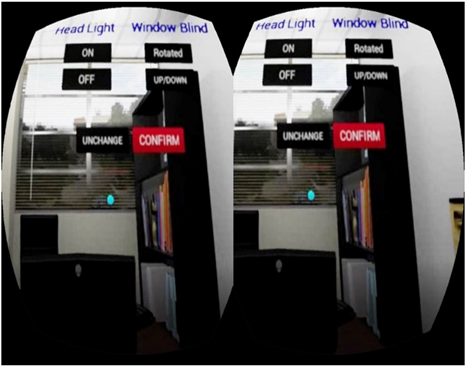}}
	\caption{The Virtual Single-Occupancy Office.}
	\label{fig2}
\end{figure}

Data of occupant behaviors of new designs are retrieved from a previous study \cite{saeidi2018spatial}. Saeidi et al. \cite{saeidi2018spatial} used IVE to study occupant light switching behaviors in a virtual single-occupancy office as shown in Fig.~\ref{fig2}. The IVE experiments were setup by manipulating critical events of the data obtained from the physical environment (e.g., arrival at the office, intermediate leaving, coming back from intermediate leaving, and departure; see TABLE~\ref{tab2}). Each event includes values of contextual factor variables (e.g., indoor and outdoor illuminance, intermediate leaving status, and occupancy status) in new-design buildings. The contextual factors (see TABLE~\ref{tab2}) were exposed to an occupant in event based experiments. The occupant’s interactions with the light switch were captured. For instance, the occupant switched the light on when indoor and outdoor were dark. A total of 180 data points relating to occupant preferences (lighting) and values of contextual factor variables (indoor and outdoor illuminance, intermediate leaving status, and occupancy status) were acquired from the IVE experiments; 36 initial events before arrival at the office, 36 events of arrival at the office, 18 events of intermediate short leave, 18 events of returning from the intermediate short leave, 18 events of intermediate long leave, 18 events of returning from the intermediate long leave, and 36 events of departure. 

Due to small sample size of the IVE data and the fact that the experiment is sequence-events, data augmentation are performed. A Hidden Markov Model (HMM) Baum-Welch algorithm \cite{baum1972inequality} is trained on the data obtained from the IVE experiment which is then used to generate synthetic samples IID. 

In the HMM, the hidden states and the observations of events are classified. The status of the light switch are classified as the hidden states. The statuses of the other variables, namely occupancy status, intermediate leaving, outdoor illuminance, and working area illuminance are classified as observations. Each observation vector is encoded to an ordinal variable by combining statuses of factors. For instance, non-occupancy, short intermediate leaving, bright outdoor illuminance, bright work area illuminance are combined as “no + short + bright + bright” and encoded by using a single value such as “1”. The transition and observation probabilities are calculated based on obtained IVE data. The HMM learns the relationship between the hidden states and observations from the transition and observation probabilities. After training process finishes, the IID synthetic sequence of events and observations (the IID synthetic IVE dataset) are randomly synthesized through the trained HMM \cite{chokwitthaya2017enhancing}.

\begin{table}[htbp]
	\caption{Statuses of Factors}
	\renewcommand\arraystretch{1.2}
	\begin{center}
		\centering\begin{tabular}{|c|>{\centering\arraybackslash}m{4cm}|c|>{\centering\arraybackslash}m{3cm}|}

			\hline
			{\textbf{Contextual Factor}} & {\textbf{Status}} \\[0.05cm]\cline{1-2}

			\multirow{2}{*}{\textbf{Occupancy}} & \multicolumn{1}{c|}{Non-Occupancy} \\[0.05cm]\cline{2-2}
												& \multicolumn{1}{c|}{Occupancy} \\[0.05cm]\cline{1-2}

			\multirow{3}{*}{\textbf{Outdoor Illuminance}} 	& \multicolumn{1}{c|}{Dark (200 Lux)} \\[0.05cm]\cline{2-2}
														  	& \multicolumn{1}{c|}{Normal (500 Lux)} \\[0.05cm]\cline{2-2}
														  	& \multicolumn{1}{c|}{Bright (700 Lux)} \\[0.05cm]\cline{1-2}
														  	
			\multirow{3}{*}{\textbf{Intermediate Leaving}}	& \multicolumn{1}{c|}{None} \\[0.05cm]\cline{2-2}
			& \multicolumn{1}{c|}{Short leaving } \\[0.05cm]\cline{2-2}
			& \multicolumn{1}{c|}{Long leaving } \\[0.05cm]\cline{1-2}

			\textbf{Independent Variable} & {\textbf{Status}} \\[0.05cm]\cline{1-2}

			\multirow{3}{*}{\textbf{Work Area Illuminance}} & \multicolumn{1}{c|}{Dark (200 Lux)} \\[0.05cm]\cline{2-2}
			& \multicolumn{1}{c|}{Normal (500 Lux)} \\[0.05cm]\cline{2-2}
			& \multicolumn{1}{c|}{Bright (700 Lux)} \\[0.05cm]\cline{1-2}
			
		\end{tabular}
		\label{tab2}
	\end{center}
\end{table}

\subsection{Generative Adversarial Network (GAN)}

\subsubsection{Data Organization}

Since the existing BPM and the target datasets have only working area illuminance as an independent variable, the missing data for contextual factors in the existing BPM and the target datasets, e.g., occupancy and intermediate leaving statuses are randomly generated from those of the synthetic IVE dataset. For instance, since occupancy status in the synthetic IVE dataset include non-occupancy and occupancy, the data of occupancy in the existing BPM dataset are randomly generated with non-occupancy and occupancy. Corresponding to the status of intermediate leaving in the synthetic IVE dataset, the data for intermediate leaving in the existing BPM are randomly generated with none, short, and long leaving.

\subsubsection{Computation}

In both experiments, we provide the generator (G) using an existing BPM and the synthetic IVE datasets (z) as input. The existing BPM and the synthetic IVE datasets are combined by concatenating. The generator is an ANN consisting of a three-layer perceptron network that has one input and two hidden layers followed by  an output layer. The inputs in the input layer are the occupancy status, intermediate leaving, and working area illuminance. The output in the output layer is the probability of switching the light on. The hidden layers of the network comprise 300 hidden neurons each with rectified linear unit activation function (ReLU) since it has been shown to have better fitting ability than the sigmoid function in similar applications \cite{sun2014deep}. To prevent overfitting, elastic net regularization (combination of L1 (Laplacian) and L2 (Gaussian) penalties) is used \cite{zou2005regularization}. The sigmoid activation function is applied at the output neuron because the outputs are probabilities. The loss function of the model is binary cross entropy (logistic regression). The learning rate and regularization are 10$^{-6}$.

The discriminator (D) is an ANN used to discriminate the outputs from the generator and the performance targets. The discriminator comprises of a three-layer ANN that has one input and two hidden layers followed by an  output layer. The setup of the discriminator is similar to the generator except that the activation functions at the hidden layers are Leaky ReLUs. Two datasets, i.e., the output of the generator and the targets are combined by concatenating. The labels of the two datasets are defined as 0 (the output of the generator) and 1 (the performance targets). 

Based on \cite{goodfellow2014generative}, learning a generator distribution $p_g$ over the performance target ($x$),  entails the generator  learning a function  from the joint distribution of the existing BPM dataset and synthetic IVE dataset  $p_{z}(z)$ to the generator output space $G(z; $$\theta_g$$)$ \cite{goodfellow2014generative}. The data space of the discriminator $D(x; $$\theta_d$$)$ will output the probability that $x$ follows the performance target distribution ($p_{data}$) instead of  $p_g$ \cite{goodfellow2014generative}. Based on \cite{goodfellow2014generative}, we train G and D together using backpropagation that minimizes $log(1-D(G(z))) + log\:D(x)$ \cite{goodfellow2014generative}. This is equivalent to playing a minimax game between G and D that has the value function $V(D,G)$ \cite{goodfellow2014generative}. The combinations of the two datasets were used as the input and the labels were used as the outputs in the discriminator.  

\begin{figure*}[htbp]
	\centerline{\includegraphics[width=0.8\paperwidth]{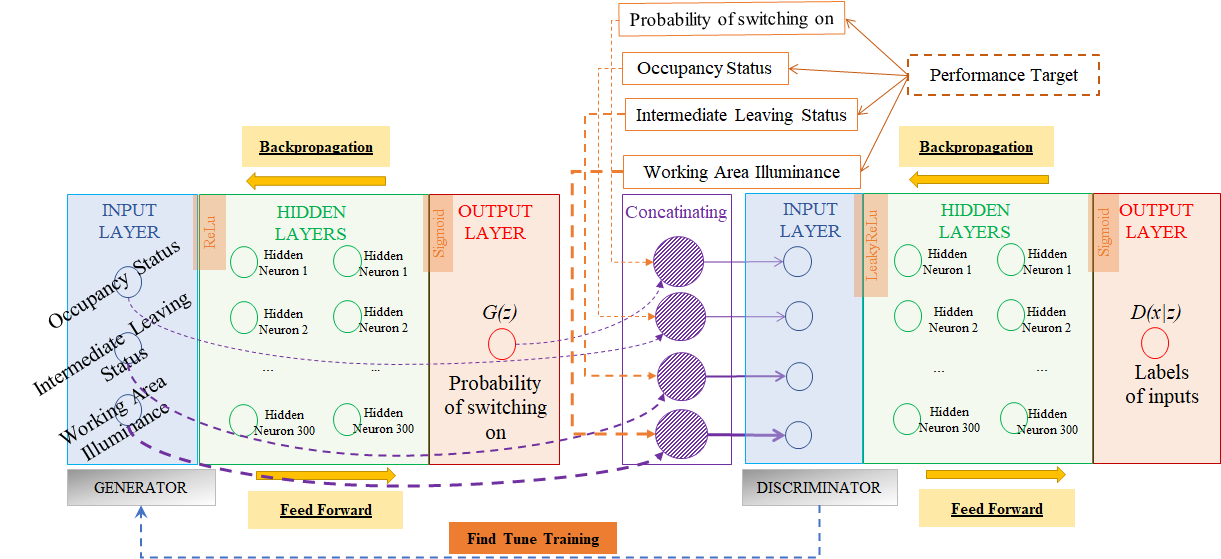}}
	\caption{Scheme of GAN of The Case Study.}
	\label{fig3}
\end{figure*}

If we use traditional GANs, the discriminator is confronted with the difficulty of accurately discriminating outputs of the generator and the targets since there is only one feature (probability of switch the light on) as the input for the discriminator. To solve the problem, we partially adapt the concept of conditional GANS \cite{mirza2014conditional} by using information of input features of the generator (occupancy status, intermediate leaving status, and working area illuminance) as additional inputs to the discriminator model. The scheme of GAN of the computational framework is shown in Fig.~\ref{fig3}. Therefore, the value function $V(D,G)$ becomes \cite{goodfellow2014generative, mirza2014conditional}: 

\begin{equation}
\begin{aligned}
\mathop{min}_{G}\mathop{max}_{D}V(D,G) = {} & \mathbb{E}_{x\sim p_{data}(x)}[logD(x|z)]+ \\
										 	& \mathbb{E}_{z\sim p_{z}(z)} [log(1-D(G(z)))]
\end{aligned}
\label{eq2}
\end{equation}  

For clarity, we summarize the corresponding pseudo-code of the optimization algorithm of the computational framework in Algorithm~\ref{algo} \cite{goodfellow2014generative, mirza2014conditional}.

\begin{algorithm}
\caption{The Optimization of The Framework. All experiments in the paper used the default values $\alpha$ = r = 10$^{-6}$, m = 2000, n = $2\mathrm{e}{5}$.}
\label{algo}

\begin{algorithmic}[1]
\Require $\alpha$, the learning rate. r, regularization. m, the batch size. n, the number of epochs.

\For{n}
\State \textbf{Train the discriminator}
\State Sample batch of $2m$ samples, ($z_{(1)}$, $\dotsc$,$z_{(2m)}$), from the generator distribution $p_{g}(z)$. To make additional inputs in the discriminator, samples $(z)$ include inputs of the generator.   
\State Sample batch of $2m$ samples from performance target,$p_{targets}(x)$.
\State Train the discriminator by using backpropagation with stochastic gradient ascent \cite{goodfellow2014generative, mirza2014conditional}:
\begin{fleqn}[\dimexpr\leftmargini-\labelsep]
	\begin{equation}
	\nabla_{\theta_{d}} \frac{1}{2m} \sum\limits_{i=1}^{2m} [logD(x^{(i)}|z^{(i)}) + log(1-D(G(z^{(i)})))]
	\end{equation}
\end{fleqn}%

\State \textbf{Train the generator}
\State Sample batch of m samples from existing BPMs dataset.
\State Sample batch of m samples from Synthetic IVE dataset.
\State Combine samples of existing BPM dataset and IVE dataset by concatenating, ($z_{(1)}$, $\dotsc$,$z_{(2m)}$).
\State Train the generator by using backpropagation with stochastic gradient descent \cite{goodfellow2014generative, mirza2014conditional}:
\begin{fleqn}[\dimexpr\leftmargini-\labelsep]
	\begin{equation}
	\nabla_{\theta_{d}} \frac{1}{2m} \sum\limits_{i=1}^{2m} [log(1-D(G(z^{(i)})))]
	\end{equation}
\end{fleqn}%

\EndFor

\end{algorithmic}
\end{algorithm}

\section{Results}
\subsection{Comparisions of Performance of BPMs}
The probabilities of switching on are randomly sampled from augmented BPMs, existing BPMs, synthetic IVE datasets, and the performance target. The probabilities of switching on are compared among three models. The mean absolute errors (MAEs) are used to determine the performance of each BPM against targets by using \eqref{eq3}. 

\begin{equation}
MAE = \frac{\sum\limits_{i=1}^{n} \abs{y_{i}-x_{i}}}{n}
\label{eq3}
\end{equation}

\newenvironment{conditions*}
{\par\vspace{\abovedisplayskip}\noindent
	\tabularx{\columnwidth}{>{$}l<{$} @{}>{${}}c<{{}$}@{} >{\raggedright\arraybackslash}X}}
{\endtabularx\par\vspace{\belowdisplayskip}}
where:
\begin{conditions*}
	i     	& = &  ranges over the list of data points, i.e., work area illuminances $(i = 1, 2, 3,\dotsc, n)$, \\
	y_{i}   & = &  refers to the probability of switching on at data point $i$ as specified by the performance targets, \\   
	x_{i} 	& = &  refers to the probability of switching on at data point i of the augmented BPM (resp. existing BPM, resp. synthetic IVE dataset).
\end{conditions*}

\begin{figure*}[htbp]
	\centering
	\begin{adjustbox}{width=\textwidth}
		\begin{subfigure}{0.4\textwidth}
			\centering
			\begin{tikzpicture}
			\begin{axis}[width = 6.0cm, height = 5.5cm,
			xmin=200, xmax=700, ymin=-0.05,ymax=1.05,
			width=\textwidth,
			legend style={at={(0.6,0.15)},anchor=south,draw=none,fill=none,column sep=10pt, nodes={scale=0.8, transform shape}},
			legend cell align={left},
			y label style={at={(axis description cs:-0.18,0.5)},anchor=north},
			x label style={at={(axis description cs:0.5,-0.08)},anchor=north},
			xlabel = \textbf{Work area illuminance (Lux)},
			ylabel = \textbf{Probability of switching on}   ]
			
			\addplot +[only marks,mark=square*,mark size=2pt] table[col sep=comma, x = W1, y = Pe1]{data1.txt};
			\addlegendentry{Performance Target};
			
			\addplot +[only marks,mark=triangle*,mark size=2pt] table[col sep=comma, x = W1, y = Au1]{data1.txt};
			\addlegendentry{Augmented BPM};
			
			\addplot +[only marks,mark=otimes*,mark size=2pt] table[col sep=comma, x = W1, y = Ex1]{data1.txt};
			\addlegendentry{Existing BPM};
			
			\addplot +[only marks,mark=diamond*,mark size=2pt] table[col sep=comma, x = W1, y = Sy1]{data1.txt};
			\addlegendentry{Synthetic IVE};
			
			\end{axis}
			\end{tikzpicture}
		\caption{\centering Experiment 1}
		\label{fig41}
		\end{subfigure}
		
		\begin{subfigure}{0.4\textwidth}
			\centering
			\begin{tikzpicture}
			\begin{axis}[width = 6.0cm, height = 5.5cm,
			xmin=200, xmax=700, ymin=-0.05,ymax=1.05,
			width=\textwidth,
			y label style={at={(axis description cs:-0.18,0.5)},anchor=north},
			x label style={at={(axis description cs:0.5,-0.08)},anchor=north},
			legend style={at={(0.6,0.15)},anchor=south,draw=none,fill=none,column sep=10pt, nodes={scale=0.8, transform shape}},
			legend cell align={left},
			xlabel = \textbf{Work area illuminance (Lux)},
			ylabel = \textbf{Probability of switching on}   ]
			
			\addplot +[only marks,mark=square*,mark=square*,mark size=2pt] table[col sep=comma, x = W2, y = Pe2]{data1.txt};
			\addlegendentry{Performance Target};
			
			\addplot +[only marks,mark=triangle*,mark size=2pt] table[col sep=comma, x = W2, y = Au2]{data1.txt};
			\addlegendentry{Augmented BPM};
			
			\addplot +[only marks,mark=otimes*,mark size=2pt] table[col sep=comma, x = W2, y = Ex2]{data1.txt};
			\addlegendentry{Existing BPM};
			
			\addplot +[only marks,mark=diamond*,mark size=2pt] table[col sep=comma, x = W2, y = Sy2]{data1.txt};
			\addlegendentry{Synthetic IVE};
			\end{axis}
			\end{tikzpicture}
		\caption{\centering Experiment 2}
		\label{fig42}
		\end{subfigure}
	\end{adjustbox}
\caption{Result of Experiments.}
\label{fig4}
\end{figure*}
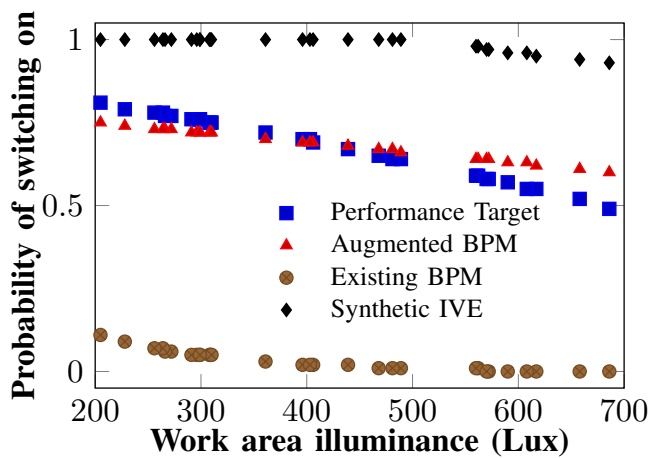
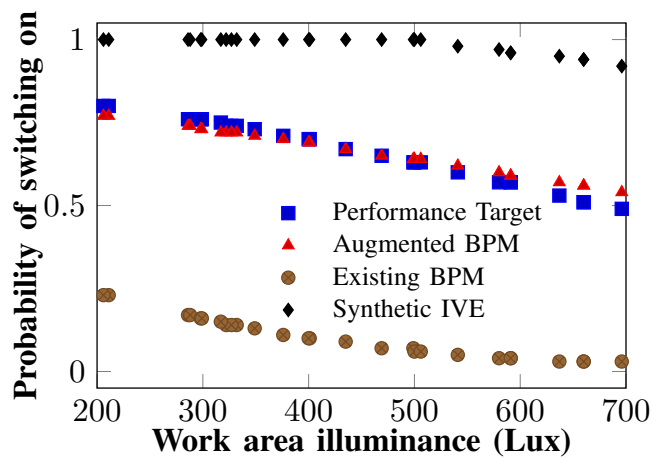

The results of the experiments are plotted in Fig.~\ref{fig4}\subref{fig41} and Fig.~\ref{fig4}\subref{fig42} to visually distinguish the performance of BPMs. The MAEs are shown in TABLE~\ref{tab3a}. From TABLE~\ref{tab3a}, the MAEs measured between probabilities of switching on as predicted by the augmented BPMs and that specified by the performance target are smallest compared to that predicted by the existing BPM or acquired from the synthetic IVE data in both experiments. The results can be interpreted as evidence that the augmented BPMs outperform both existing BPMs and IVEs.

\begin{table}[htbp]
	\caption{Results of MAEs}
	\renewcommand\arraystretch{1.1}
	\begin{center}
		\begin{tabular}{|>{\centering\arraybackslash}m{0.5cm}|>{\centering\arraybackslash}m{0.5cm}|>{\centering\arraybackslash}m{0.5cm}|>{\centering\arraybackslash}m{0.5cm}|>{\centering\arraybackslash}m{0.5cm}|>{\centering\arraybackslash}m{0.5cm}|>{\centering\arraybackslash}m{0.5cm}|}
			
			\hline
			\multirow{2}{*}{} & \multicolumn{3}{c|}{\parbox[c]{0.1\textwidth}{\centering \textbf{Experiment 1}}} & \multicolumn{3}{c|}{\parbox[c]{0.1\textwidth}{\centering \textbf{Experiment 2}}}\\[0.01cm]\cline{2-7}
				
				& {\rotatebox[origin=c]{90}{\parbox[c]{1.5cm}{\centering \textbf{Augmented BPM}}}} &	
				{\rotatebox[origin=c]{90}{\parbox[c]{1.5cm}{\centering \textbf{Existing BPM}}}} &
				{\rotatebox[origin=c]{90}{\parbox[c]{1.5cm}{\centering \textbf{Synthetic IVE}}}} &
				{\rotatebox[origin=c]{90}{\parbox[c]{1.5cm}{\centering \textbf{Augmented BPM}}}} &	
				{\rotatebox[origin=c]{90}{\parbox[c]{1.5cm}{\centering \textbf{Existing BPM}}}} &
				{\rotatebox[origin=c]{90}{\parbox[c]{1.5cm}{\centering \textbf{Synthetic IVE}}}}\\\cline{2-7}
				
			\hline
			\multicolumn{1}{|c|}{\textbf{MAE}} & 0.17 & 0.48 & 0.47 & 0.14 & 0.41 & 0.47 \\[0.01cm]
			\hline
			
		\end{tabular}
		\label{tab3a}
	\end{center}
\end{table}

\subsection{Tests of The Performance of Augmented BPMs}
To show that the predictions obtained from the augmented BPMs produced by the computational framework outperform the that obtained from existing BPMs and the probabilities acquired from the synthetic IVE dataset, we apply statistical analysis to find significant difference of errors measured between: 1) the performance targets and the existing BPMs, and 2) the performance targets and the synthetic IVE dataset, and 3) the performance targets and the augmented BPMs.

The performance of the existing BPMs, the IVE, and the augmented BPMs are investigated by using absolute errors as measured values as shown in TABLE~\ref{tab4}. 

\begin{table}[htbp]
	\caption{Comparison of Performance of BPMs}
	\renewcommand\arraystretch{1.0}
	\begin{center}
		\begin{tabular}{|>{\centering\arraybackslash}m{1cm}|>{\centering\arraybackslash}m{6cm}|}
			\hline
			{\textbf{Absolute error}} & {\textbf{Explanation}}\\[0.01cm]
			
			\hline
			$E_{1}$ & $|$probability of switching the light on obtained from the existing BPM $-$ the performance target$|$\\[0.01cm]
			
			\hline
			$E_{2}$ & $|$probability of switching the light on obtained from the IVE $-$ the performance target$|$\\[0.01cm]
			
			\hline
			$E_{3}$ & $|$probability of switching the light on obtained from the augmented BPM $-$ the performance target$|$\\[0.01cm]
			
			\hline			
		\end{tabular}
		\label{tab4}
	\end{center}
\end{table}

To statistically test the significance of the performances of augmented BPMs for both experiments, hypotheses are defined as follows:
 
To test the performance of the augmented BPMs and the existing BPMs, hypothesis 1 is defined as follow:

\begin{center} 

$H_{0}$: mean of $E_{1}$ $-$ mean of $E_{3}$ $=$ 0

$H_{1}$: mean of $E_{1}$ $-$ mean of $E_{3}$ $>$ 0

\end{center}

To test the performance of the augmented BPMs and the IVE, hypothesis 2 is defined as follow:

\begin{center}

\centering $H_{0}$: mean of $E_{2}$ $-$ mean of $E_{3}$ $=$ 0

\centering $H_{1}$: mean of $E_{2}$ $-$ mean of $E_{3}$ $>$ 0

\end{center}

A one tailed t-test ($\alpha$ = 0.05) was applied to investigate statistically significant difference between the performance of the augmented BPMs, and the existing BPMs as well as the IVE. The results are shown in TABLE~\ref{tab5}.

\begin{table}[htbp]
	\caption{Results of the Hypothesis Testing}
	\renewcommand\arraystretch{1.1}
	\begin{center}
		\centering\begin{tabular}{|*{5}{c|}}
			\hline
			\multirow{3}{*}{} 	& \multicolumn{2}{c|}{\textbf{Experiment 1}} & \multicolumn{2}{c|}{\textbf{Experiment 2}}\\[0.01cm]\cline{2-5} 
								
								& \multicolumn{2}{c|}{$Hypothesis$} & \multicolumn{2}{c|}{$Hypothesis$} \\[0.01cm] \cline{2-5}
			
								& \multicolumn{1}{c|}{\textit{1}} & \multicolumn{1}{c|}{\textit{2}} & \multicolumn{1}{c|}{\textit{1}} & \multicolumn{1}{c|}{\textit{2}} \\[0.01cm]
			\hline
			{\parbox{0.06\textwidth}{\centering \textbf{Absolute T-value}}} & \multicolumn{1}{c|}{44.300} & \multicolumn{1}{c|}{17.873} & \multicolumn{1}{c|}{53.535} & \multicolumn{1}{c|}{19.377} \\[0.02cm]
			
			\hline
			{\parbox{0.06\textwidth}{\centering \textbf{P-value}}} & \multicolumn{1}{c|}{$<$ 0.05} & \multicolumn{1}{c|}{$<$ 0.05} & \multicolumn{1}{c|}{$<$ 0.05} & \multicolumn{1}{c|}{$<$ 0.05} \\[0.01cm]
			
			\hline
			{\parbox{0.06\textwidth}{\centering {\textbf{\textit{$H_{0}$}}}}} & \multicolumn{1}{c|}{Reject} & \multicolumn{1}{c|}{Reject} & \multicolumn{1}{c|}{Reject} & \multicolumn{1}{c|}{Reject} \\[0.01cm]
			\hline
			
		\end{tabular}
		\label{tab5}
	\end{center}
\end{table}

From TABLE~\ref{tab5}, the null hypotheses were rejected for all cases. Based on the hypotheses testing, we conclude that, the probabilities of switching the light on estimated by the augmented BPM are significantly closer to the performance targets than that estimated by the existing BPMs or the (synthetic) IVE dataset. This shows a strong potential of using the computational framework to enhance performance of BPMs and reduce performance discrepancy between prediction during designs and operational buildings.  

\section{Conclusion}
The paper presents a computation framework to reduce the performance discrepancy between predictions during designs and the actual performance observed when building is in operation. GANs are used to learn a mixture model that allows appropriate combination of existing BPMs with knowledge of occupant behaviors responding to contextual factors in new designs as obtained from IVE experiments. 

The results of the experiments show promising potential of the computational framework for reducing the performance discrepancy. From the evidence in TABLE~\ref{tab5}, the augmented BPMs from both experiments outperform existing BPMs and IVEs.  
  
In the future work, uncertainties have to be considered to improve the performance of the framework. There are many factors that may contribute to uncertainties such as quality of IVE datasets, existing BPMs, and the system of the framework. More IVE experiments need to be conducted to investigate and improve the performances of IVEs in occupant behavior study and enhance accuracy of the framework. Furthermore, the quality of IVE datasets may be dependent on many elements such as cues, instrument, and occupants. Study of cues may need to be explored to enhance the quality of IVE datasets. Since the data of existing BPMs are obtained from occupant behaviors in existing buildings, specified constraints on types and behaviors of occupants may need to be defined corresponding to occupant in new design. The algorithm of the framework may need to be further improved to increase accuracy of augmented BPMs.   

\section*{Acknowledgment}
This paper was partially supported by the U.S. National Science Foundation Award \#1640818. Any opinions, findings, and conclusions or recommendations expressed in this material are those of the author(s) and do not necessarily reflect the views of the Louisiana Board of Regents or the National Science Foundation.



\bibliographystyle{IEEEtran}

\begin{thebibliography}{10}
\providecommand{\url}[1]{#1}
\csname url@samestyle\endcsname
\providecommand{\newblock}{\relax}
\providecommand{\bibinfo}[2]{#2}
\providecommand{\BIBentrySTDinterwordspacing}{\spaceskip=0pt\relax}
\providecommand{\BIBentryALTinterwordstretchfactor}{4}
\providecommand{\BIBentryALTinterwordspacing}{\spaceskip=\fontdimen2\font plus
\BIBentryALTinterwordstretchfactor\fontdimen3\font minus
  \fontdimen4\font\relax}
\providecommand{\BIBforeignlanguage}[2]{{%
\expandafter\ifx\csname l@#1\endcsname\relax
\typeout{** WARNING: IEEEtran.bst: No hyphenation pattern has been}%
\typeout{** loaded for the language `#1'. Using the pattern for}%
\typeout{** the default language instead.}%
\else
\language=\csname l@#1\endcsname
\fi
#2}}
\providecommand{\BIBdecl}{\relax}
\BIBdecl

\bibitem{van2016review}
C.~Van~Dronkelaar, M.~Dowson, E.~Burman, C.~Spataru, and D.~Mumovic, ``A review
  of the energy performance gap and its underlying causes in non-domestic
  buildings,'' \emph{Frontiers in Mechanical Engineering}, vol.~1, p.~17, 2016.

\bibitem{frei2017performance}
B.~Frei, C.~Sagerschnig, and D.~Gyalistras, ``Performance gaps in swiss
  buildings: an analysis of conflicting objectives and mitigation strategies,''
  \emph{Energy Procedia}, vol. 122, pp. 421--426, 2017.

\bibitem{hong2016occupant}
T.~Hong, H.~Sun, Y.~Chen, S.~C. Taylor-Lange, and D.~Yan, ``An occupant
  behavior modeling tool for co-simulation,'' \emph{Energy and Buildings}, vol.
  117, pp. 272--281, 2016.

\bibitem{o2014contextual}
W.~O'Brien and H.~B. Gunay, ``The contextual factors contributing to occupants'
  adaptive comfort behaviors in offices--a review and proposed modeling
  framework,'' \emph{Building and Environment}, vol.~77, pp. 77--87, 2014.

\bibitem{attia2012development}
S.~Attia, A.~Evrard, and E.~Gratia, ``Development of benchmark models for the
  egyptian residential buildings sector,'' \emph{Applied Energy}, vol.~94, pp.
  270--284, 2012.

\bibitem{feng2016preliminary}
X.~Feng, D.~Yan, C.~Wang, and H.~Sun, ``A preliminary research on the
  derivation of typical occupant behavior based on large-scale questionnaire
  surveys,'' \emph{Energy and Buildings}, vol. 117, pp. 332--340, 2016.

\bibitem{santin2009effect}
O.~G. Santin, L.~Itard, and H.~Visscher, ``The effect of occupancy and building
  characteristics on energy use for space and water heating in dutch
  residential stock,'' \emph{Energy and buildings}, vol.~41, no.~11, pp.
  1223--1232, 2009.

\bibitem{andersen2011modelling}
R.~V. Andersen, B.~W. Olesen, and J.~Toftum, ``Modelling occupants’ heating
  set-point prefferences,'' in \emph{Building Simulation 2011: 12th Conference
  of International Building Performance Simulation Association}, 2011.

\bibitem{hunt1980predicting}
D.~Hunt, ``Predicting artificial lighting use-a method based upon observed
  patterns of behaviour,'' \emph{Lighting Research \& Technology}, vol.~12,
  no.~1, pp. 7--14, 1980.

\bibitem{kinateder2014social}
M.~Kinateder, M.~M{\"u}ller, M.~Jost, A.~M{\"u}hlberger, and P.~Pauli, ``Social
  influence in a virtual tunnel fire--influence of conflicting information on
  evacuation behavior,'' \emph{Applied ergonomics}, vol.~45, no.~6, pp.
  1649--1659, 2014.

\bibitem{kobes2010exit}
M.~Kobes, I.~Helsloot, B.~de~Vries, and J.~Post, ``Exit choice,(pre-) movement
  time and (pre-) evacuation behaviour in hotel fire evacuation—behavioural
  analysis and validation of the use of serious gaming in experimental
  research,'' \emph{Procedia Engineering}, vol.~3, pp. 37--51, 2010.

\bibitem{osman2015impact}
O.~A. Osman, J.~Codjoe, and S.~Ishak, ``Impact of time-to-collision information
  on driving behavior in connected vehicle environments using a driving
  simulator test bed,'' \emph{Traffic Logist. Eng}, vol.~3, no.~1, 2015.

\bibitem{rumschlag2015effects}
G.~Rumschlag, T.~Palumbo, A.~Martin, D.~Head, R.~George, and R.~L. Commissaris,
  ``The effects of texting on driving performance in a driving simulator: The
  influence of driver age,'' \emph{Accident Analysis \& Prevention}, vol.~74,
  pp. 145--149, 2015.

\bibitem{heydarian2014immersive}
A.~Heydarian, J.~P. Carneiro, D.~Gerber, B.~Becerik-Gerber, T.~Hayes, and
  W.~Wood, ``Immersive virtual environments: experiments on impacting design
  and human building interaction,'' 2014.

\bibitem{niu2016virtual}
S.~Niu, W.~Pan, and Y.~Zhao, ``A virtual reality integrated design approach to
  improving occupancy information integrity for closing the building energy
  performance gap,'' \emph{Sustainable cities and society}, vol.~27, pp.
  275--286, 2016.

\bibitem{nicol2001characterising}
J.~F. Nicol, ``Characterising occupant behaviour in buildings: towards a
  stochastic model of occupant use of windows, lights, blinds, heaters and
  fans,'' in \emph{Proceedings of the seventh international IBPSA conference,
  Rio}, vol.~2, 2001, pp. 1073--1078.

\bibitem{newsham1994manual}
G.~Newsham, ``Manual control of window blinds and electric lighting:
  implications for comfort and energy consumption,'' \emph{Indoor Environment},
  vol.~3, no.~3, pp. 135--144, 1994.

\bibitem{reinhart2004lightswitch}
C.~F. Reinhart, ``Lightswitch-2002: a model for manual and automated control of
  electric lighting and blinds,'' \emph{Solar energy}, vol.~77, no.~1, pp.
  15--28, 2004.

\bibitem{reinhart2003lightswitch}
C.~Reinhart, M.~Morrison, and F.~Dubrous, ``The lightswitch wizard-reliable
  daylight simulations for initial design investigation,'' in \emph{8th
  International IBPSA Conference, Eindhoven, The Netherlands}, vol.~3, 2003,
  pp. 1093--1100.

\bibitem{bourgeois2004adding}
D.~Bourgeois, J.~Hand, I.~Mcdonald, and C.~Reinhart, ``Adding sub-hourly
  occupancy prediction, occupancy-sensing control and manual environmental
  control to esp-r,'' \emph{Proceeding of Esim 2004, Vancouver, BC}, pp.
  119--126, 2004.

\bibitem{gunay2017development}
H.~B. Gunay, W.~O'Brien, I.~Beausoleil-Morrison, and S.~Gilani, ``Development
  and implementation of an adaptive lighting and blinds control algorithm,''
  \emph{Building and Environment}, vol. 113, pp. 185--199, 2017.

\bibitem{da2013occupants}
P.~C. da~Silva, V.~Leal, and M.~Andersen, ``Occupants interaction with electric
  lighting and shading systems in real single-occupied offices: Results from a
  monitoring campaign,'' \emph{Building and Environment}, vol.~64, pp.
  152--168, 2013.

\bibitem{boyce1980observations}
P.~Boyce, ``Observations of the manual switching of lighting,'' \emph{Lighting
  Research \& Technology}, vol.~12, no.~4, pp. 195--205, 1980.

\bibitem{rijal2007using}
H.~B. Rijal, P.~Tuohy, M.~A. Humphreys, J.~F. Nicol, A.~Samuel, and J.~Clarke,
  ``Using results from field surveys to predict the effect of open windows on
  thermal comfort and energy use in buildings,'' \emph{Energy and buildings},
  vol.~39, no.~7, pp. 823--836, 2007.

\bibitem{sadeghi2016occupant}
S.~A. Sadeghi, P.~Karava, I.~Konstantzos, and A.~Tzempelikos, ``Occupant
  interactions with shading and lighting systems using different control
  interfaces: A pilot field study,'' \emph{Building and Environment}, vol.~97,
  pp. 177--195, 2016.

\bibitem{gunay2016modeling}
H.~B. Gunay, W.~O’Brien, I.~Beausoleil-Morrison, and S.~Gilani, ``Modeling
  plug-in equipment load patterns in private office spaces,'' \emph{Energy and
  Buildings}, vol. 121, pp. 234--249, 2016.

\bibitem{heydarian2015immersive}
A.~Heydarian, J.~P. Carneiro, D.~Gerber, and B.~Becerik-Gerber, ``Immersive
  virtual environments, understanding the impact of design features and
  occupant choice upon lighting for building performance,'' \emph{Building and
  Environment}, vol.~89, pp. 217--228, 2015.

\bibitem{saeidi2015measuring}
S.~Saeidi, T.~Rizzuto, Y.~Zhu, and R.~Kooima, ``Measuring the effectiveness of
  an immersive virtual environment for the modeling and prediction of occupant
  behavior,'' in \emph{Sustainable Human--Building Ecosystems}, 2015, pp.
  159--167.

\bibitem{saeidi2018spatial}
S.~Saeidi, C.~Chokwitthaya, Y.~Zhu, and M.~Sun, ``Spatial-temporal event-driven
  modeling for occupant behavior studies using immersive virtual
  environments,'' \emph{Automation in Construction}, vol.~94, pp. 371--382,
  2018.

\bibitem{lecun2015deep}
Y.~LeCun, Y.~Bengio, and G.~Hinton, ``Deep learning,'' \emph{nature}, vol. 521,
  no. 7553, p. 436, 2015.

\bibitem{goodfellow2014generative}
I.~Goodfellow, J.~Pouget-Abadie, M.~Mirza, B.~Xu, D.~Warde-Farley, S.~Ozair,
  A.~Courville, and Y.~Bengio, ``Generative adversarial nets,'' in
  \emph{Advances in neural information processing systems}, 2014, pp.
  2672--2680.

\bibitem{Collier2}
E.~Collier, R.~DiBiano, and S.~Mukhopadhyay, ``Cactusnets: Layer applicability
  as a metric for transfer learning,'' in \emph{2018 International Joint
  Conference on Neural Networks, {IJCNN} 2018, Rio de Janeiro, Brazil, July
  8-13, 2018}, 2018, pp. 1--8.

\bibitem{Saikat}
S.~Basu, S.~Mukhopadhyay, M.~Karki, R.~DiBiano, S.~Ganguly, R.~R. Nemani, and
  S.~Gayaka, ``Deep neural networks for texture classification - {A}
  theoretical analysis,'' \emph{Neural Networks}, vol.~97, pp. 173--182, 2018.

\bibitem{DeepSAT}
S.~Basu, S.~Ganguly, S.~Mukhopadhyay, R.~DiBiano, M.~Karki, and R.~R. Nemani,
  ``Deepsat: a learning framework for satellite imagery,'' in \emph{Proceedings
  of the 23rd {SIGSPATIAL} International Conference on Advances in Geographic
  Information Systems, Bellevue, WA, USA, November 3-6, 2015}, 2015, pp.
  37:1--37:10.

\bibitem{collier}
E.~Collier, K.~Duffy, S.~Ganguly, G.~Madanguit, S.~Kalia, S.~Gayaka, R.~R.
  Nemani, A.~R. Michaelis, S.~Li, A.~R. Ganguly, and S.~Mukhopadhyay,
  ``Progressively growing generative adversarial networks for high resolution
  semantic segmentation of satellite images,'' in \emph{2018 {IEEE}
  International Conference on Data Mining Workshops, {ICDM} Workshops,
  Singapore, Singapore, November 17-20, 2018}, 2018, pp. 763--769.

\bibitem{ledig2017photo}
C.~Ledig, L.~Theis, F.~Husz{\'a}r, J.~Caballero, A.~Cunningham, A.~Acosta,
  A.~Aitken, A.~Tejani, J.~Totz, Z.~Wang \emph{et~al.}, ``Photo-realistic
  single image super-resolution using a generative adversarial network,'' in
  \emph{Proceedings of the IEEE conference on computer vision and pattern
  recognition}, 2017, pp. 4681--4690.

\bibitem{radford2015unsupervised}
A.~Radford, L.~Metz, and S.~Chintala, ``Unsupervised representation learning
  with deep convolutional generative adversarial networks,'' \emph{arXiv
  preprint arXiv:1511.06434}, 2015.

\bibitem{wang2016generative}
X.~Wang and A.~Gupta, ``Generative image modeling using style and structure
  adversarial networks,'' in \emph{European Conference on Computer
  Vision}.\hskip 1em plus 0.5em minus 0.4em\relax Springer, 2016, pp. 318--335.

\bibitem{wu2016learning}
J.~Wu, C.~Zhang, T.~Xue, B.~Freeman, and J.~Tenenbaum, ``Learning a
  probabilistic latent space of object shapes via 3d generative-adversarial
  modeling,'' in \emph{Advances in neural information processing systems},
  2016, pp. 82--90.

\bibitem{baum1972inequality}
L.~Baum, ``An inequality and associated maximization technique in statistical
  estimation of probabilistic functions of a markov process,''
  \emph{Inequalities}, vol.~3, pp. 1--8, 1972.

\bibitem{chokwitthaya2017enhancing}
C.~Chokwitthaya, R.~Dibiano, S.~Saeidi, S.~Mukhopadhyay, and Y.~Zhu,
  ``Enhancing the prediction of artificial lighting control behavior using
  virtual reality (vr): A pilot study,'' in \emph{Construction Research
  Congress 2018}, 2017, pp. 216--223.

\bibitem{sun2014deep}
Y.~Sun, X.~Wang, and X.~Tang, ``Deep learning face representation from
  predicting 10,000 classes,'' in \emph{Proceedings of the IEEE conference on
  computer vision and pattern recognition}, 2014, pp. 1891--1898.

\bibitem{zou2005regularization}
H.~Zou and T.~Hastie, ``Regularization and variable selection via the elastic
  net,'' \emph{Journal of the royal statistical society: series B (statistical
  methodology)}, vol.~67, no.~2, pp. 301--320, 2005.

\bibitem{mirza2014conditional}
M.~Mirza and S.~Osindero, ``Conditional generative adversarial nets,''
  \emph{arXiv preprint arXiv:1411.1784}, 2014.

\end{thebibliography}

\end{document}